\documentclass[11pt]{article}
\usepackage[margin=1in]{geometry}
\usepackage[T1]{fontenc}
\usepackage{lmodern}
\usepackage{microtype}
\usepackage{amsmath,amssymb}
\usepackage{booktabs}
\usepackage{graphicx}
\usepackage{xcolor}
\usepackage{hyperref}
\usepackage[numbers,sort&compress]{natbib}
\usepackage{caption}
\usepackage{enumitem}
\hypersetup{colorlinks=true,linkcolor=teal!70!black,citecolor=teal!70!black,urlcolor=teal!70!black,pdfauthor={Jithin VG, Ditto PS}}
\setlist{itemsep=2pt,topsep=4pt}
\newcommand{\pp}{\,pp}

\title{Bud ContextCompact 2.0\\[0.4em]\large Context Compaction for Agent Success and Accuracy with Small Language Models and Constrained Context Windows}
\author{Jithin VG\\Bud Ecosystem\\\texttt{jithinvg@bud.studio}
  \and
  Ditto PS\\Bud Ecosystem\\\texttt{dittops@bud.studio}}
\date{October 2026}

\begin{document}
\maketitle

\begin{abstract}
Agents that work for many turns outgrow the context window of the model that serves them, and on small language models (SLMs) and constrained inference systems that window is small. Context compaction replaces older turns with a summary so a session can continue, but it is lossy, it costs a model call, and whether it helps or hurts accuracy depends on when it runs. We describe Bud ContextCompact 2.0, the compaction built into the Bud agent runtime, and evaluate it in two settings. On recall questions over recorded agent sessions, compaction to a 16K budget raised a 1.2B-parameter model's accuracy from 0.247 (truncated history) and 0.176 (full 32K history) to 0.335 ($p<0.001$ against the full history), and compaction to a 32K budget raised a 9B model from 0.515 (truncated) to 0.577 ($p=0.04$). On agentic software engineering, 100 SWE-bench Verified tasks run end to end by a 27B reasoning model on a 49K-token window (556 graded sessions), compaction raised the resolve rate on long sessions from 41\% to 53\% (+12.3\pp, 95\% CI $+1.6$ to $+23.0$, exact McNemar $p=0.043$), left short sessions unchanged (+1.7\pp, $p=1.0$), and did so at equal total token cost and 13\% fewer tokens per solved task. The dominant failure without compaction was not the history overflowing but a reasoning turn exhausting the space the history left. Two conditions bound the gain: under a 45-minute session budget it vanished ($-3.3$\pp, $p=0.48$), because each summary took a median 161\,s; and in a threshold study it appeared only when compaction triggered early enough to leave room (24K threshold, $+13.9$\pp{} on long pairs; 48K, compaction ran in 4 of 50 sessions and the arm behaved like no compaction). An 8B model on the same tasks submitted after about ten turns at 5--8K tokens and never needed compaction. We argue that the accuracy results transfer across hardware for an equivalent model and window, while the time cost scales with the hardware and must be budgeted for.
\end{abstract}

\section{Introduction}
\label{sec:intro}

Tool-using agents alternate model calls and tool calls until a task is done~\cite{yao2023react}. Every observation, command output and intermediate thought is appended to the conversation, so the prompt grows with every turn. Software-engineering agents are an extreme case: a single SWE-bench task~\cite{jimenez2024swebench,openai2024verified} routinely takes tens to hundreds of turns, many of them reading files and test logs~\cite{yang2024sweagent,wang2025openhands}. When the conversation no longer fits the model's context window, the session fails.

The window available to an agent depends on where it runs. Hosted frontier models offer windows of hundreds of thousands of tokens. Small language models (SLMs) deployed on a single GPU, a desktop accelerator or a CPU usually run with 8K to 64K: the KV cache that a longer window needs competes for scarce memory, prefill time grows with the prompt, and many small models degrade well before their nominal limit~\cite{liu2024lost,hsieh2024ruler}. These are exactly the deployments where a long-running agent hits the wall first.

Context compaction keeps a session inside its window by replacing older turns with a summary~\cite{packer2023memgpt}. It is not free. The summary is lossy; writing it costs a model call; it can invalidate the provider's prefix cache; and it can run at the wrong time---too late to prevent an overflow, or so often that the agent spends its time summarising. Whether compaction improves an agent's success, and under which conditions, is an empirical question that published agent evaluations rarely isolate, because they change scaffolding, prompts and models at the same time.

This paper isolates it. We evaluate Bud ContextCompact 2.0, the compaction built into the Bud agent runtime (Section~\ref{sec:system}), with A/B comparisons in which the compaction setting is the only difference between arms. We report:
\begin{itemize}
  \item \textbf{SLM evidence on recall} (Section~\ref{sec:prior}): with 1.2B and 9B models answering questions over recorded agent sessions, compaction to a constrained budget beat both truncation and, for the 1.2B model, the full history.
  \item \textbf{Agentic evidence} (Section~\ref{sec:swe}): on 100 SWE-bench Verified tasks run by a 27B reasoning model, compaction raised the resolve rate on long sessions by 12.3 points ($p=0.043$), with a replication on 40 tasks added after the first result.
  \item \textbf{Mechanism} (Section~\ref{sec:mechanism}): with a reasoning model, the binding constraint was the room left for a turn's reasoning, not the history itself.
  \item \textbf{Boundary conditions} (Sections~\ref{sec:budget}--\ref{sec:size}): the gain vanishes under a tight wall-clock budget, depends on the trigger threshold, and needs sessions long enough to compact---which an 8B model on these tasks never produced.
  \item \textbf{Cost} (Section~\ref{sec:cost}): equal total tokens, 13\% fewer tokens per solved task, every summary at full measured fidelity.
\end{itemize}
Section~\ref{sec:discussion} discusses what transfers to SLM and CPU deployments and what does not.

\section{Background and related work}
\label{sec:related}

\paragraph{Agents and long contexts.} ReAct-style agents interleave reasoning and tool use~\cite{yao2023react}. Software-engineering agents such as SWE-agent~\cite{yang2024sweagent} and OpenHands~\cite{wang2025openhands} run for many turns on SWE-bench~\cite{jimenez2024swebench}; SWE-bench Verified is a human-validated 500-task subset~\cite{openai2024verified}. Customer-service benchmarks $\tau$-bench and $\tau^2$-bench~\cite{yao2024taubench,barres2025tau2} test tool-agent-user interaction over long dialogues. Models use long contexts unevenly: evidence in the middle of a long prompt is used worse than evidence at either end~\cite{liu2024lost}, and effective context is often far shorter than the advertised window~\cite{hsieh2024ruler}.

\paragraph{Managing the context.} MemGPT pages information between a bounded working context and external memory under model control~\cite{packer2023memgpt}. Prompt-compression methods such as LLMLingua and LongLLMLingua drop low-information tokens before a call~\cite{jiang2023llmlingua,jiang2024longllmlingua}. Agent scaffolds commonly truncate old observations or summarise the history when it grows. These approaches are usually evaluated on retrieval or question answering, or as one component of a scaffold whose other parts change at the same time. We evaluate compaction alone, inside a fixed agent, on task success.

\paragraph{Serving.} Our models are served by vLLM~\cite{kwon2023vllm}. Extending a model's window beyond its training length uses rope scaling such as YaRN~\cite{peng2024yarn}. The models are from the Qwen3 family~\cite{yang2025qwen3}.

\section{Bud ContextCompact 2.0}
\label{sec:system}

ContextCompact runs inside the Bud agent runtime (\texttt{budprompt}), which owns each conversation: a client sends only the new items of a turn and links it to the previous response, and the runtime rebuilds the history from its store. Compaction is an agent-level setting, \texttt{context\_management = \{compaction: off|auto, compact\_threshold, fold\_style\}}, so two versions of one agent can differ in compaction alone.

\paragraph{Budget.} For a window $W$ the runtime reserves $R$ tokens for the reply (the request's output limit or reasoning budget, else 4,096, capped at $W/4$) and a 512-token margin for the serving framework's message framing, giving an input budget $B = W - R - 512$. The compaction threshold $T$ is the agent's setting, either in tokens or as a share of $B$, clamped to $[0.3B,\,0.95B]$ (the lower bound was $0.5B$ before the threshold study of Section~\ref{sec:threshold}). The window comes from the deployment record: for a self-hosted model, the serving engine's \texttt{max-model-len}, never above the model's own limit.

\paragraph{When it runs.} When a turn's prompt would pass $T$, the runtime starts a \emph{background} compaction and serves the turn with the full history; the summary is applied from the next turn on. If a turn would not fit the window at all, the runtime compacts \emph{inline} before the model call; and if the model call itself overflows mid-run, an in-run rescue compacts and retries. A compaction is stored as a checkpoint on the last history item it covers; later turns replay the checkpoint's summary followed by the items after it, so the stored history is never modified and successive compactions (\emph{generations}) chain.

\paragraph{What it writes.} The runtime calls BudCache's stateless compaction service, which selects the span to fold and asks a writer model---in our deployment the agent's own model, through the same inference gateway---for a \emph{briefing}: one plain-language checkpoint covering the goal, labelled facts and identifiers, work done, work pending, current work and the next step, written from the previous briefing plus every message folded since. Each briefing is checked before use: critical literals (paths, identifiers, error lines, numbers) are matched against the source, and literals the writer cites that occur in no source are removed. A background briefing that fails the fidelity check is discarded rather than applied.

\paragraph{Relation to 1.x.} The earlier, standalone ContextCompact service (Section~\ref{sec:prior}) keeps per-session state and adds tool-output masking with recoverable stubs, retrieval of earlier evidence into a footer before the question, and model-written ``gist'' folds alongside deterministic literal channels. Version 2.0 moves compaction into the runtime that owns the conversation and uses the stateless briefing path, so any agent on the platform gets it by setting one field.

\section{Evidence from ContextCompact 1.x: small models and tight windows}
\label{sec:prior}

Before the runtime integration, the standalone ContextCompact service was evaluated on small models and constrained windows. We summarise the results that bear on this paper's question.

\subsection{Recall over recorded agent sessions}

\paragraph{Setup.} 42 recorded sessions---Claude Code working on a large Rust repository, $\tau$-bench customer service~\cite{yao2024taubench}, tool-use and WildChat conversations---with 358 questions whose answers were planted in the history and validated against an oracle. Each session is replayed turn by turn through the compactor, so masking, folds and recall happen as in production; each question is then asked as the next user message, and answers are graded by substring match. Three arms are answered by the same model: \emph{full}, the unlimited history (or the model's real window for local models); \emph{bound}, the history truncated to the compactor's budget, keeping the system prompt and the task message---what an agent without compaction must send in the same window; and \emph{compacted}. Each small model answers and writes its own folds, served by LM Studio on a single NVIDIA GB10 desktop system. These questions test whether information survives, not whether a task is completed.

\begin{figure}[t]
  \centering
  \includegraphics[width=0.85\linewidth]{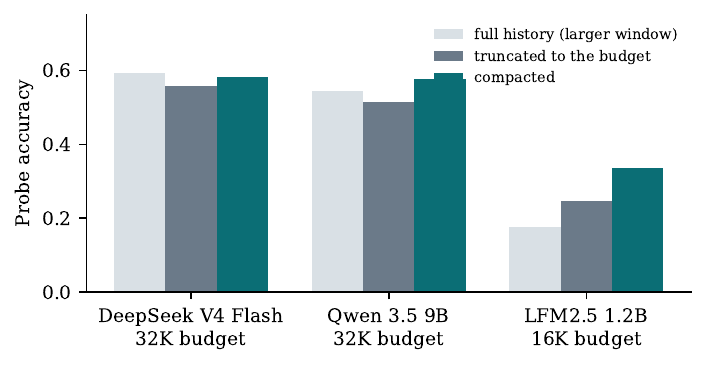}
  \caption{Recall-probe accuracy (ContextCompact 1.x). DeepSeek: 358 probes, mean of three runs. Small models: 239 probes, one run each, served locally on an NVIDIA GB10. The full-history arm uses a larger window than the compacted arm has: 1M for DeepSeek, 64K for Qwen 3.5 9B, 32K for LFM2.5 1.2B.}
  \label{fig:slm}
\end{figure}

\begin{table}[t]
  \centering
  \footnotesize
  \begin{tabular}{llrrrll}
    \toprule
    Model & Budget & Full & Bound & Compacted & vs full & vs bound \\
    \midrule
    DeepSeek V4 Flash & 32K & 0.592 & 0.559 & 0.581 & n.s.\ ($p\ge0.27$) & $+$, n.s.\ ($p\ge0.13$) \\
    Qwen 3.5 9B & 32K (full: 64K) & 0.544 & 0.515 & \textbf{0.577} & $+8$, $p=0.32$ & $+15$, $p=0.04$ \\
    LFM2.5 1.2B & 16K (full: 32K) & 0.176 & 0.247 & \textbf{0.335} & $+38$, $p<0.001$ & $+21$, $p=0.006$ \\
    \bottomrule
  \end{tabular}
  \caption{Recall-probe accuracy and net answers gained (paired McNemar). DeepSeek: per-run $p$ over three runs; small models: one run each. Compacted prompts were 0.52 (DeepSeek), 0.34 (9B) and 0.40 (1.2B) of the full history's tokens.}
  \label{tab:slm}
\end{table}

\paragraph{Results.} For the large model, compaction to a 32K budget cost no detectable accuracy against the unlimited history at 52\% of the tokens, and beat truncation in every run without reaching significance in any (Table~\ref{tab:slm}). For small models compaction was an accuracy gain rather than only a cost saving. Qwen 3.5 9B beat truncation by 15 net answers ($p=0.04$) and, on the Claude Code sessions, also beat its own 64K full history ($+16$, $p=0.009$), while on $\tau$-bench it lost 8 net answers to the 64K history ($p=0.08$) and tied truncation. LFM2.5 1.2B compacted to 16K beat both truncation ($+21$, $p=0.006$) and its full 32K history ($+38$, $p<0.001$), with a third of the tokens---although its own writer produced a usable summary in only 13 of 62 attempts, the rest falling back to deterministic skeleton folds. Small models degrade with context length far more than large ones, so a short view with the needed evidence next to the question helps them outright.

Two caveats. The small-model results are single runs on 239 probes; DeepSeek's identical runs varied by about $\pm0.005$ overall. And an earlier harness layout, which put the answering instruction into the question and hence into the recall query, moved every arm (the full arm from about 0.63 to 0.59) and gave the 1.2B model only parity with truncation (0.251 against 0.255); only numbers from one layout are comparable.

\subsection{Agentic tasks in tight windows}

The 1.x service was also run end to end on $\tau^2$-bench~\cite{barres2025tau2} with Bonsai-27B and Qwen 3.5 9B, in windows of 10K to 16K tokens, against a different baseline: the automatic compaction of the DeepSeek Harness (DSH), which trims large tool results and then summarises the oldest span into a model-written checkpoint, re-summarising earlier checkpoints as it goes. These runs compare two compaction strategies, not compaction against none, and their results are mixed:
\begin{itemize}
  \item \textbf{Reasoning on, tight windows.} At 14K, ContextCompact solved 141 tasks to DSH's 122 pooled over two 100-task rounds ($+16\%$, $p\approx0.02$); at 12K, 36 to 28 ($p=0.096$). At 16K, three replicates of 100 tasks moved in both directions (79/73, 81/71, 70/76), a wash.
  \item \textbf{Reasoning off, 10K.} DSH was ahead (35 to 23; 32 to 22, $p=0.013$) until 1.x adopted briefing summaries with protection of live tool outputs, which brought it to parity (29 to 30) at 29\% lower cost; at 12K that configuration solved 41 against DSH's 35 ($p=0.18$) at 42\% lower cost per solved task.
  \item \textbf{Qwen 3.5 9B} tied DSH at 16K (90/89) and 14K (85/85).
  \item \textbf{Mechanism.} In a controlled replay of 24 recorded loop points, a briefing raised the share of samples that made progress on the next step from 26\% to 47\% ($p=0.012$).
\end{itemize}
For reference, with DeepSeek V4 Flash on $\tau^2$ at 10K, sending the raw history until it overflowed solved 0.45 of tasks, a sliding-window truncation 0.75, and the compacting arms 0.85--0.94 (18--20 tasks per arm, not significant). These results are why the 2.0 runtime defaults to briefing summaries, and they show that the choice of summary strategy matters as much as compacting at all, especially for models that do not reason.

\section{Agentic study: SWE-bench Verified}
\label{sec:swe}

\subsection{Setup}

\paragraph{Agent and arms.} One Bud agent with four tools (\texttt{bash}, \texttt{view}, \texttt{edit}, \texttt{submit}) and a fixed system prompt, in two versions that differ only in \texttt{context\_management}: \textbf{off} (no compaction) and \textbf{auto} (compaction with $T=0.5B$, briefing summaries). The tools are executed by the client in each task's official SWE-bench container; every turn is a \texttt{/v1/responses} call that sends only the new tool outputs and links to the previous response, so the runtime holds the conversation and decides when to compact. A session ends when the agent calls \texttt{submit}, when a turn fails (for example by exhausting the window), or at a limit; its patch is then graded by the official SWE-bench harness on the hidden FAIL\_TO\_PASS and PASS\_TO\_PASS tests.

\paragraph{Model and serving.} Qwen3.8-27B with thinking enabled, temperature 0.6, top-$p$ 0.95, parallel tool calls, and no output cap, served by vLLM on four NVIDIA H100 80GB GPUs (one replica each) with a 49,429-token window and prefix caching disabled. With no output cap the reserve is the 4,096-token default, so $B=44{,}821$ and $T=22{,}410$ tokens. The summariser inherits the agent's reasoning settings; its deadline was 600\,s.

\paragraph{Tasks.} 100 SWE-bench Verified tasks in two batches. The first 60 are 12 pilot tasks plus 48 sampled with a fixed seed, stratified 18/26/16 across the dataset's ``under 15 minutes'', ``15 minutes--1 hour'' and ``1--4 hours'' difficulty labels; hard tasks are oversampled on purpose because compaction can only matter in long sessions. Because session length in the first batch tracked the repository far more than the difficulty label (django tasks went long in 16 of 54 pairs, astropy in 6 of 6, sympy in 17 of 26), the 40 added tasks were sampled with a fixed seed from non-django tasks labelled 15 minutes or longer. Each task runs twice per arm.

\paragraph{Conditions.} \emph{Budgeted}: 60 turns or 45 minutes per session, on the first 60 tasks. \emph{No budget}: 250 turns and no clock, on the 39 first-batch tasks where any budgeted session hit a limit and on all 40 added tasks; the 21 first-batch tasks that never approached a limit keep their budgeted sessions, whose outcomes are already unconstrained. No session in the no-budget condition reached 250 turns.

\paragraph{Analysis.} Sessions are paired on (task, repeat). We test the difference in resolution with the exact McNemar test on discordant pairs~\cite{mcnemar1947}, and report the mean difference with a 95\% bootstrap interval that resamples tasks (10,000 resamples), since the two repeats of a task are not independent~\cite{efron1994bootstrap}. Single-arm rates carry Wilson intervals~\cite{wilson1927}. A pair is \emph{long} when either arm's prompt crossed $T$ at some turn: the only pairs in which compaction could act. The split uses both arms alike; splitting on whether the auto session compacted would condition on the treated arm's own trajectory and is not used for inference. The split and the test were fixed before the 40 added tasks ran.

\subsection{Accuracy}

\begin{figure}[t]
  \centering
  \includegraphics[width=0.85\linewidth]{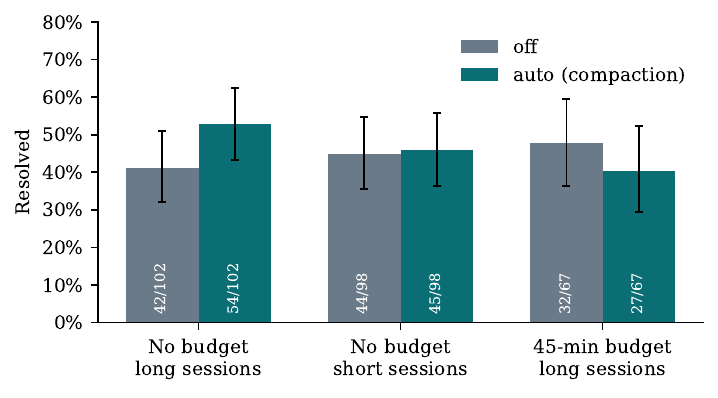}
  \caption{Resolve rate by arm with Wilson 95\% intervals. Long pairs: either arm's prompt crossed the compaction threshold. Without a session budget, compaction raised the resolve rate on long sessions and left short ones unchanged; under a 45-minute budget the gain vanished.}
  \label{fig:accuracy}
\end{figure}

\begin{table}[t]
  \centering
  \small
  \begin{tabular}{lrrrrrr}
    \toprule
    No budget & Pairs & Off & Auto & Only off & Only auto & Auto$-$off (95\% CI), $p$ \\
    \midrule
    Long, first 60 tasks & 60 & 27 & 35 & 3 & 11 & $+13.5$ ($+2.7$, $+25.7$), 0.057 \\
    Long, 40 added tasks & 42 & 15 & 19 & 6 & 10 & $+10.4$ ($-10.4$, $+31.2$), 0.45 \\
    \textbf{Long, all 100 tasks} & \textbf{102} & \textbf{42} & \textbf{54} & \textbf{9} & \textbf{21} & $\mathbf{+12.3}$ ($+1.6$, $+23.0$), \textbf{0.043} \\
    Short, all 100 tasks & 98 & 44 & 45 & 4 & 5 & $+1.7$ ($-5.1$, $+8.5$), 1.0 \\
    All pairs, 100 tasks & 200 & 86 & 99 & 13 & 26 & $+6.5$ ($+0.5$, $+13.0$), 0.053 \\
    \bottomrule
  \end{tabular}
  \caption{Paired resolution without a session budget. Differences in percentage points; intervals from a task-level bootstrap; $p$ from the exact McNemar test.}
  \label{tab:accuracy}
\end{table}

Without a session budget, compaction raised the resolve rate on long sessions from 41.2\% to 52.9\% (Table~\ref{tab:accuracy}, Figure~\ref{fig:accuracy}). Among long pairs that only one arm resolved, compaction won 21 to 9 ($p=0.043$), and the task-level bootstrap interval excludes zero. The 40 added tasks, run after the first result to test it, moved the same way and by a similar amount ($+10.4$ against $+13.5$ points) but are not significant on their own; the pooled result depends on both batches. On short sessions, where compaction never ran, the arms are indistinguishable (4 against 5 discordant pairs), as they should be if the arms differ only in compaction. Reweighted to SWE-bench Verified's difficulty distribution, the first batch's no-budget resolve rates are 60.5\% (off) and 66.4\% (auto); absolute rates are otherwise not comparable to leaderboard figures, since the sample favours hard tasks and the agent scaffold is minimal.

\subsection{Mechanism: reasoning needs room}
\label{sec:mechanism}

\begin{figure}[t]
  \centering
  \begin{minipage}[t]{0.58\linewidth}\centering
    \includegraphics[width=\linewidth]{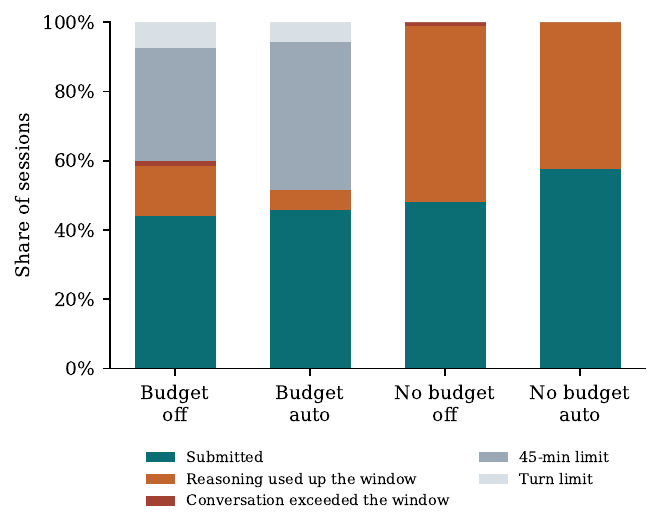}
    \caption{How sessions ended. Without a budget, most failures are a single reasoning turn exhausting the window.}
    \label{fig:endings}
  \end{minipage}\hfill
  \begin{minipage}[t]{0.40\linewidth}\centering
    \includegraphics[width=\linewidth]{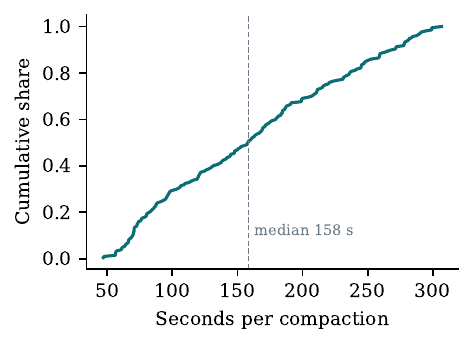}
    \caption{Time per compaction, all 207 checkpoints written during the study (median 158\,s).}
    \label{fig:time}
  \end{minipage}
\end{figure}

\begin{figure}[t]
  \centering
  \includegraphics[width=0.8\linewidth]{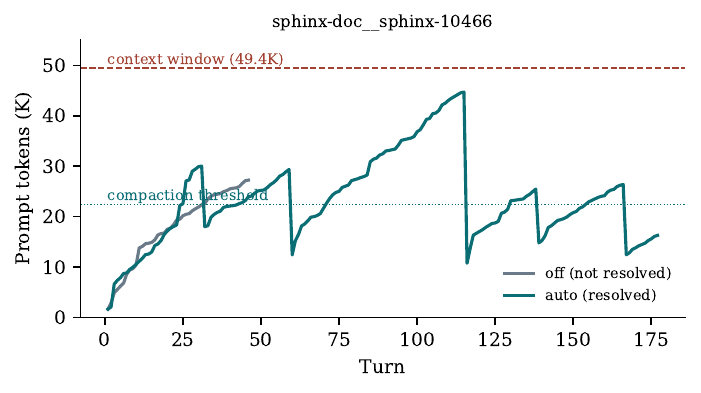}
  \caption{Prompt size per turn for one task and repeat. Each drop is a compaction taking effect. Background compaction applies from the turn after it finishes, so the prompt can run past the threshold before it drops.}
  \label{fig:trace}
\end{figure}

Without a budget, 102 of 200 uncompacted sessions ended because one turn's reasoning used up the rest of the window, against 85 with compaction; only 2 uncompacted sessions ended because the conversation itself outgrew the window (Figure~\ref{fig:endings}). The history was not the problem at the moment of failure: the prompt at the last successful turn had a median of 17.8K tokens without compaction and 13.5K with it, against a 49K window. The model falls into a long reasoning loop and spends whatever room the history leaves. Compaction keeps the history short, which leaves that loop more room to finish, and 115 compacted sessions reached \texttt{submit} against 96 uncompacted. Figure~\ref{fig:trace} shows one task: the uncompacted session failed mid-reasoning at 27K tokens after 46 turns, while the compacted session compacted five times and resolved the task after 177 turns.

\subsection{Under a time budget the gain vanishes}
\label{sec:budget}

With 60 turns or 45 minutes per session, the arms resolved 63/120 (off) and 59/120 (auto) on the first 60 tasks ($-3.3$\pp, 95\% CI $-10.8$ to $+4.2$, $p=0.48$), and on long pairs 32/67 against 27/67 ($p=0.30$). Auto sessions hit the 45-minute limit more often (51 against 39). Each compaction took a median of 161\,s (p90 263\,s, Figure~\ref{fig:time}) because the summariser runs on the agent's own reasoning model; background compaction overlaps this with the agent's turns, but the session still pays for the summariser's share of the shared GPUs and for the turns it keeps working. Compaction converts ``ran out of room'' into ``kept working'', which helps only if the session is allowed to keep working.

\subsection{The threshold must leave room}
\label{sec:threshold}

\begin{figure}[t]
  \centering
  \includegraphics[width=0.6\linewidth]{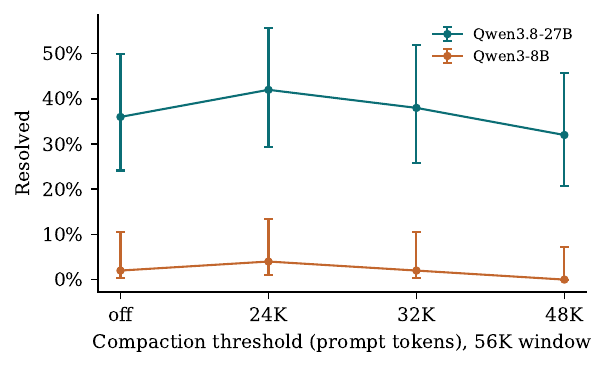}
  \caption{Resolve rate by compaction threshold on a 56,320-token window (25 tasks, 2 repeats, 50 sessions per point, Wilson intervals).}
  \label{fig:threshold}
\end{figure}

\begin{table}[t]
  \centering
  \small
  \begin{tabular}{lrrrr}
    \toprule
    Qwen3.8-27B, 56K window & Off & 24K & 32K & 48K \\
    \midrule
    Resolved (of 50) & 18 & 21 & 19 & 16 \\
    Sessions that compacted & -- & 20 & 16 & 4 \\
    Reasoning exhausted the window & 29 & 26 & 24 & 31 \\
    Long pairs: only off / only threshold & & 2 / 7 & 3 / 3 & 6 / 6 \\
    Long pairs: threshold $-$ off (pp) & & $+13.9$ & $0.0$ & $+2.4$ \\
    Long pairs: McNemar $p$ & & 0.18 & 1.0 & 1.0 \\
    \bottomrule
  \end{tabular}
  \caption{Threshold study. Long pairs: either arm's prompt crossed 24K (33--36 pairs per threshold).}
  \label{tab:threshold}
\end{table}

To see how the trigger point matters we lowered the runtime's threshold floor to $0.3B$, moved the 27B deployment to a 56,320-token window ($B=51{,}712$), and ran 25 tasks that had run long in the main study, twice each, with compaction off or at 24K, 32K or 48K tokens (Table~\ref{tab:threshold}, Figure~\ref{fig:threshold}). The effect follows how often each threshold compacts at all. At 24K compaction ran in 20 of 50 sessions, and on long pairs it won 7 to 2 (+13.9 points), close to the main study's +12.3 at a 22.4K threshold. At 48K it ran in only 4 sessions: most sessions ended when a reasoning turn exhausted the window before the conversation reached 48K, so the arm behaved like no compaction. With 25 tasks none of these differences is significant; the trend is consistent with the main result rather than independent evidence for it.

\subsection{Model size: an 8B model never needed compaction}
\label{sec:size}

We repeated the threshold study with Qwen3-8B, extended to the same 56,320-token window with YaRN rope scaling (factor 2)~\cite{peng2024yarn}. The 8B model resolved 1, 2, 1 and 0 of 50 sessions in the four arms. It called \texttt{submit} after a median of about ten turns, at a peak prompt of 5--8K tokens, and almost half its submissions were empty patches. Compaction ran once in 200 sessions, so the four arms were the same agent in practice. On these tasks the 8B model's limitation is not context; it stops before the context grows. This bounds the claim of Section~\ref{sec:swe} to models capable enough to work long, and is why the SLM evidence in this paper comes from recall tasks over long recorded sessions (Section~\ref{sec:prior}), where the context is long by construction.

\subsection{Cost and compaction behaviour}
\label{sec:cost}

\begin{table}[t]
  \centering
  \small
  \begin{tabular}{lrr}
    \toprule
    No budget, 100 tasks, 200 sessions per arm & Off & Auto \\
    \midrule
    Prompt tokens & 98.9M & 95.0M \\
    Output tokens & 9.54M & 11.30M \\
    Summariser tokens (in / out) & -- & 1.51M / 0.94M \\
    All tokens, summariser included & 108.4M & 108.7M \\
    \textbf{All tokens per resolved session} & \textbf{1.26M} & \textbf{1.10M} \\
    Prompt tokens a prefix cache could reuse & 98\% & 96\% \\
    Model time & 73.8\,h & 84.2\,h \\
    Session time (median) & 23.8\,min & 22.3\,min \\
    \bottomrule
  \end{tabular}
  \caption{Cost. The prefix-cache row is modelled from the traces: a turn reuses the previous turn's prompt and output unless a compaction rewrote the history. Reasoning tokens are included in output tokens.}
  \label{tab:cost}
\end{table}

Compaction did not make sessions cheaper overall: including the summariser, the arms used the same tokens (Table~\ref{tab:cost}), and the compacted arm used 14\% more model time because it kept working on tasks the uncompacted arm abandoned when its window ran out (on the 40 added tasks alone, 11\% more tokens and 42\% more model time). Per solved task it was 13\% cheaper. Because each compaction rewrites the history, it costs one turn of prefix-cache reuse: modelled from the traces, 96\% of compacted prompt tokens could have come from a prefix cache, against 98\% uncompacted.

The 134 compactions applied in the study (75 sessions, 132 in the background) all passed the fidelity check: mean critical-literal recall 1.000, no invented literals, no summariser errors, no fallbacks. Each reduced the conversation from a mean of 24.7K to 11.4K tokens ($-54\%$) with a 2.3K-token briefing; one session compacted seven times in succession. Nineteen further checkpoints finished after their session's last turn and never took effect.

\section{Discussion}
\label{sec:discussion}

\paragraph{What transfers to SLM and CPU deployments.} For a given model, weights, window, sampling configuration and task, the tokens an agent produces do not depend on the hardware that computes them, apart from small numerical differences between kernels. The accuracy results of Sections~\ref{sec:prior} and~\ref{sec:swe} therefore describe the same configurations served on a CPU or an edge accelerator. The mechanism of Section~\ref{sec:mechanism} also points where such deployments sit: the gain comes from leaving a reasoning turn room in a window that is tight relative to how much the model reasons, and constrained deployments run exactly such windows. The recall study's largest gains (Figure~\ref{fig:slm}) are on its smallest model and smallest budget.

\paragraph{What does not transfer: time.} Hardware does not change which tokens are produced, but it changes how long they take, and Section~\ref{sec:budget} shows that time can change the accuracy result. A summary that took a median 161\,s on an H100 with a 27B reasoning writer takes far longer on a CPU. Any deployment with a time budget---per session, per user wait, or per batch---pays that penalty in proportion. For constrained systems this argues for a writer that is cheaper than the agent: reasoning disabled for the summariser, a smaller writer model, or the extractive fold style, which writes without a model call. We did not measure these on agentic tasks.

\paragraph{When to compact.} Compaction helped only when it ran early enough to leave room (Section~\ref{sec:threshold}): a threshold near the window edge fires after the failure it was meant to prevent. Background compaction applies from the following turn, so the room left at the threshold must cover at least one more turn's tool output and reasoning. For reasoning models in tight windows, a threshold around half the input budget worked. The runtime's default of $0.8B$ (about 36K tokens on the 49K window) lies between the 32K and 48K arms of Section~\ref{sec:threshold}, which compacted in 16 and 4 of 50 sessions respectively; for reasoning agents in tight windows a lower setting is the safer choice.

\section{Conclusion}

Context compaction is not a free lunch, but on the right workload it is a measurable gain. For a reasoning agent working long in a window that is tight relative to its reasoning, ContextCompact 2.0 raised the resolve rate on long SWE-bench Verified sessions by 12 points at no extra token cost, with every summary at full measured fidelity; on recall over recorded sessions, it lifted 1.2B and 9B models above both truncation and, for the smallest, the full history. The gain needs time to work, a threshold that fires early enough, and an agent that works long enough to need it. Those conditions, not the hardware, decide whether compaction helps.

\bibliographystyle{plainnat}
\bibliography{refs}

@inproceedings{jimenez2024swebench,
  title     = {{SWE}-bench: Can Language Models Resolve Real-World {GitHub} Issues?},
  author    = {Jimenez, Carlos E. and Yang, John and Wettig, Alexander and Yao, Shunyu and Pei, Kexin and Press, Ofir and Narasimhan, Karthik},
  booktitle = {International Conference on Learning Representations (ICLR)},
  year      = {2024}
}

@misc{openai2024verified,
  title        = {Introducing {SWE}-bench Verified},
  author       = {{OpenAI}},
  year         = {2024},
  howpublished = {\url{https://openai.com/index/introducing-swe-bench-verified/}}
}

@inproceedings{yang2024sweagent,
  title     = {{SWE}-agent: Agent-Computer Interfaces Enable Automated Software Engineering},
  author    = {Yang, John and Jimenez, Carlos E. and Wettig, Alexander and Lieret, Kilian and Yao, Shunyu and Narasimhan, Karthik and Press, Ofir},
  booktitle = {Advances in Neural Information Processing Systems (NeurIPS)},
  year      = {2024}
}

@inproceedings{wang2025openhands,
  title     = {{OpenHands}: An Open Platform for {AI} Software Developers as Generalist Agents},
  author    = {Wang, Xingyao and others},
  booktitle = {International Conference on Learning Representations (ICLR)},
  year      = {2025}
}

@article{packer2023memgpt,
  title   = {{MemGPT}: Towards {LLMs} as Operating Systems},
  author  = {Packer, Charles and Wooders, Sarah and Lin, Kevin and Fang, Vivian and Patil, Shishir G. and Stoica, Ion and Gonzalez, Joseph E.},
  journal = {arXiv preprint arXiv:2310.08560},
  year    = {2023}
}

@article{liu2024lost,
  title   = {Lost in the Middle: How Language Models Use Long Contexts},
  author  = {Liu, Nelson F. and Lin, Kevin and Hewitt, John and Paranjape, Ashwin and Bevilacqua, Michele and Petroni, Fabio and Liang, Percy},
  journal = {Transactions of the Association for Computational Linguistics},
  volume  = {12},
  year    = {2024}
}

@inproceedings{hsieh2024ruler,
  title     = {{RULER}: What's the Real Context Size of Your Long-Context Language Models?},
  author    = {Hsieh, Cheng-Ping and Sun, Simeng and Kriman, Samuel and Acharya, Shantanu and Rekesh, Dima and Jia, Fei and Zhang, Yang and Ginsburg, Boris},
  booktitle = {Conference on Language Modeling (COLM)},
  year      = {2024}
}

@inproceedings{jiang2023llmlingua,
  title     = {{LLMLingua}: Compressing Prompts for Accelerated Inference of Large Language Models},
  author    = {Jiang, Huiqiang and Wu, Qianhui and Lin, Chin-Yew and Yang, Yuqing and Qiu, Lili},
  booktitle = {Proceedings of EMNLP},
  year      = {2023}
}

@inproceedings{jiang2024longllmlingua,
  title     = {{LongLLMLingua}: Accelerating and Enhancing {LLMs} in Long Context Scenarios via Prompt Compression},
  author    = {Jiang, Huiqiang and Wu, Qianhui and Luo, Xufang and Li, Dongsheng and Lin, Chin-Yew and Yang, Yuqing and Qiu, Lili},
  booktitle = {Proceedings of ACL},
  year      = {2024}
}

@inproceedings{yao2023react,
  title     = {{ReAct}: Synergizing Reasoning and Acting in Language Models},
  author    = {Yao, Shunyu and Zhao, Jeffrey and Yu, Dian and Du, Nan and Shafran, Izhak and Narasimhan, Karthik and Cao, Yuan},
  booktitle = {International Conference on Learning Representations (ICLR)},
  year      = {2023}
}

@article{yao2024taubench,
  title   = {$\tau$-bench: A Benchmark for Tool-Agent-User Interaction in Real-World Domains},
  author  = {Yao, Shunyu and Shinn, Noah and Razavi, Pedram and Narasimhan, Karthik},
  journal = {arXiv preprint arXiv:2406.12045},
  year    = {2024}
}

@article{barres2025tau2,
  title   = {$\tau^2$-Bench: Evaluating Conversational Agents in a Dual-Control Environment},
  author  = {Barres, Victor and Dong, Honghua and Ray, Soham and Si, Xujie and Narasimhan, Karthik},
  journal = {arXiv preprint arXiv:2506.07982},
  year    = {2025}
}

@inproceedings{kwon2023vllm,
  title     = {Efficient Memory Management for Large Language Model Serving with {PagedAttention}},
  author    = {Kwon, Woosuk and Li, Zhuohan and Zhuang, Siyuan and Sheng, Ying and Zheng, Lianmin and Yu, Cody Hao and Gonzalez, Joseph E. and Zhang, Hao and Stoica, Ion},
  booktitle = {Proceedings of the 29th Symposium on Operating Systems Principles (SOSP)},
  year      = {2023}
}

@inproceedings{peng2024yarn,
  title     = {{YaRN}: Efficient Context Window Extension of Large Language Models},
  author    = {Peng, Bowen and Quesnelle, Jeffrey and Fan, Honglu and Shippole, Enrico},
  booktitle = {International Conference on Learning Representations (ICLR)},
  year      = {2024}
}

@article{yang2025qwen3,
  title   = {Qwen3 Technical Report},
  author  = {Yang, An and others},
  journal = {arXiv preprint arXiv:2505.09388},
  year    = {2025}
}

@article{mcnemar1947,
  title   = {Note on the Sampling Error of the Difference Between Correlated Proportions or Percentages},
  author  = {McNemar, Quinn},
  journal = {Psychometrika},
  volume  = {12},
  number  = {2},
  pages   = {153--157},
  year    = {1947}
}

@article{wilson1927,
  title   = {Probable Inference, the Law of Succession, and Statistical Inference},
  author  = {Wilson, Edwin B.},
  journal = {Journal of the American Statistical Association},
  volume  = {22},
  number  = {158},
  pages   = {209--212},
  year    = {1927}
}

@book{efron1994bootstrap,
  title     = {An Introduction to the Bootstrap},
  author    = {Efron, Bradley and Tibshirani, Robert J.},
  publisher = {Chapman \& Hall/CRC},
  year      = {1994}
}

\end{document}